%% file: main.tex
\documentclass[sigconf]{acmart}

\usepackage{hyperref}
\usepackage{xcolor}

\AtBeginDocument{%
 }

\copyrightyear{2023}
\acmYear{2023}
\setcopyright{rightsretained}
\acmConference[CIKM '23]{Proceedings of the 32nd ACM International Conference on Information and Knowledge Management}{October 21--25, 2023}{Birmingham, United Kingdom}
\acmBooktitle{Proceedings of the 32nd ACM International Conference on Information and Knowledge Management (CIKM '23), October 21--25, 2023, Birmingham, United Kingdom}\acmDOI{10.1145/3583780.3615291}
\acmISBN{979-8-4007-0124-5/23/10}

\makeatletter
\gdef\@copyrightpermission{
  \begin{minipage}{0.3\columnwidth}
   \href{https://eur01.safelinks.protection.outlook.com/?url=https\%3A\%2F\%2Fcreativecommons.org\%2Flicenses\%2Fby\%2F4.0\%2F&data=05\%7C01\%7Cheydar.soudani\%40ru.nl\%7Cf49ac8384b44421ef74508dba83008f4\%7C084578d9400d4a5aa7c7e76ca47af400\%7C1\%7C0\%7C638288696492593634\%7CUnknown\%7CTWFpbGZsb3d8eyJWIjoiMC4wLjAwMDAiLCJQIjoiV2luMzIiLCJBTiI6Ik1haWwiLCJXVCI6Mn0\%3D\%7C3000\%7C\%7C\%7C&sdata=oKkScxFkoGx8yIF36OyGf7vIi1hTKqQW1wQtMFGZ7sE\%3D&reserved=0}{\includegraphics[width=0.90\textwidth]{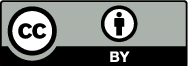}}
  \end{minipage}\hfill
  \begin{minipage}{0.7\columnwidth}
   \href{https://eur01.safelinks.protection.outlook.com/?url=https\%3A\%2F\%2Fcreativecommons.org\%2Flicenses\%2Fby\%2F4.0\%2F&data=05\%7C01\%7Cheydar.soudani\%40ru.nl\%7Cf49ac8384b44421ef74508dba83008f4\%7C084578d9400d4a5aa7c7e76ca47af400\%7C1\%7C0\%7C638288696492593634\%7CUnknown\%7CTWFpbGZsb3d8eyJWIjoiMC4wLjAwMDAiLCJQIjoiV2luMzIiLCJBTiI6Ik1haWwiLCJXVCI6Mn0\%3D\%7C3000\%7C\%7C\%7C&sdata=oKkScxFkoGx8yIF36OyGf7vIi1hTKqQW1wQtMFGZ7sE\%3D&reserved=0}{\parbox{0.99\columnwidth}{
      This work is licensed under a Creative Commons Attribution International 4.0 License.
    }}
  \end{minipage}
  \vspace{5pt}
}
\makeatother

\begin{document}

\title{Data Augmentation for Conversational AI}

\author{Heydar Soudani}
\email{heydar.soudani@ru.nl}
\affiliation{
  \institution{Radboud University}
  \city{Nijmegen}
  \country{The Netherlands}
}
\author{Evangelos Kanoulas}
\email{e.kanoulas@uva.nl}
\affiliation{
  \institution{University of Amsterdam}
  \city{Amsterdam}
  \country{The Netherlands}
}
\author{Faegheh Hasibi}
\email{f.hasibi@cs.ru.nl}
\affiliation{
  \institution{Radboud University}
  \city{Nijmegen}
  \country{The Netherlands}
}
\begin{abstract}
Advancements in conversational systems have revolutionized information access, surpassing the limitations of single queries. 
However, developing dialogue systems requires a large amount of training data, which is a challenge in low-resource domains and languages. Traditional data collection methods like crowd-sourcing are labor-intensive and time-consuming, making them ineffective in this context. Data augmentation (DA) is an affective approach to alleviate the data scarcity problem in conversational systems. This tutorial provides a comprehensive and up-to-date overview of DA approaches in the context of conversational systems. It highlights recent advances in conversation augmentation, open domain and task-oriented conversation generation, and different paradigms of evaluating these models. We also discuss current challenges and future directions in order to help researchers and practitioners to further advance the field in this area.
\end{abstract}

\keywords{Data Augmentation, Dataset Creation, Conversation Generation, Conversational AI}



\maketitle

\input{sections/introduction}
\input{sections/outline}

\input{sections/presenters}

\bibliographystyle{ACM-Reference-Format}
\bibliography{main}


\end{document}

%% file: sections/introduction.tex
\section{Introduction}

\noindent
\paragraph{\textbf{Motivation}} The development of dialogue systems has garnered significant attention and demand in both industry and everyday life due to their diverse range of functionalities that cater to user needs. These systems can be broadly classified into two categories: task-oriented dialogue systems (TOD) and open-domain dialogue systems (ODD)~\cite{Jinjie2023Recent}. TOD systems are specifically designed to address particular problems within a specific domain, with the objective of performing tasks like ticket or table reservations. Thus, their primary focus is task completion. On the other hand, ODD systems engage in unrestricted conversations on a wide range of topics~\cite{Jinjie2023Recent}. The main challenge of ODD systems thus lies in handling consistency and seamless transitioning between turns in the conversation. 

The progress of dialogue systems relies heavily on the use of large neural models, similar to other natural language processing (NLP) tasks, and as a result, their effectiveness is contingent upon the availability of substantial amounts of training data~\cite{Zamani2022CIS}.
The dependence on large scale training data challenges development of dialogue agents, particularly for low resource settings with limited or no training data.
While crowd-sourcing serves as the primary method for generating datasets in dialogue systems, it's labor intensive nature possesses limitations concerning time, cost, and scalability~\cite{li2022DIALOGIC}.
The scarcity of data for diverse and specific domains, compounded by the difficulty of adapting existing datasets or generating entirely new ones calls for alternative methods of training dialogue systems.

%

To tackle the issue of data shortage in dialogue systems, several methods have been proposed, including semi-supervised learning and data augmentation (DA)~\cite{yang2022Limited}.
While semi-supervised learning is a promising approach, relying solely on existing dialogue datasets presents a classic chicken-and-egg problem. DA, on the other hand, involves generating conversation samples from external resources, such as unstructured text files and structured data like knowledge graphs.
%
This approach serves multiple purposes: it diversifies datasets, introduces novel conversational scenarios, and enhances control over the flow of the generated conversation.

In this tutorial, we aim to offer a comprehensive overview of conversation augmentation and generation methods for TOD and ODD systems. We delve into the specific challenges that must be addressed when undertaking the task of creating new dialogue data.  To provide a comprehensive package for dialogue data creation, we also present an overview of evaluation methods and available datasets that can be utilized to assess the quality and performance of the generated dialogue data. By covering these aspects, our tutorial offers a holistic understanding of the methods, challenges, and evaluation procedures associated with the dialogue data creation.


\paragraph{\textbf{Previous tutorials.}}
Our tutorial builds upon two key concepts: conversational systems and DA. Tutorials that discuss these concepts in recent years include:

\begin{itemize}
    \item Conversational Recommender Systems~\cite{Fu2020Recommendation} in RecSys 2020.
    This tutorial focuses on the foundations and algorithms for conversational recommendation and their application in real-world systems like search engines, e-commerce, and social networks.
    
    \item Conversational Information Seeking: Theory and Application~\cite{Dalton2022Conversational} in SIGIR 2022. The tutorial aims to provide an introduction to Conversational Information Seeking (CIS), covering recent advanced topics and state-of-the-art approaches. 
    
    \item Self-Supervised Learning for Recommendation~\cite{Huang2022Self} in CIKM 2022.
    This tutorial aims to present a systematic review of state-of-the-art self-supervised learning (SSL)-based recommender systems.
    
    \item Limited Data Learning~\cite{yang2022Limited} in ACL 2022.
    This tutorial offers an overview of methods alleviating the need for labeled data in NLP, including DA and semi-supervised learning.

    \item Proactive Conversational Agents~\cite{Liao2023Proactive} in WSDM 2023. This tutorial introduces methods to equip conversational agents with the ability to interact with end users proactively.
\end{itemize}


Unlike previous tutorials that focus on either conversational systems or the data scarcity problem, this tutorial provides an in-depth exploration of the challenges associated with augmenting and creating conversational data, highlighting the unique difficulties posed in conversational context. 
To the best of our knowledge, no tutorial has specifically focused on dataset creation techniques for dialogue systems. 

\paragraph{\textbf{Target audience and prerequisites.}}
This tutorial is designed for professionals, students, and researchers in information retrieval and natural language processing, specifically interested in conversational AI.
Familiarity with machine learning, deep learning, and transformers is required. No prior knowledge of dialogue system models or data augmentation methods is necessary.


\paragraph{\textbf{Tutorial material}}
Tutorial slides, a collection of essential references, and other support materials can be found at the tutorial website \url{https://dataug-convai.github.io}.

%% file: sections/outline.tex
\section{Tutorial Outline}

We plan to give a half-day tutorial (three hours). The tutorial starts by an introduction, followed by three main sessions. We plan to have a short Q\&A after each session and conclude the tutorial with a discussion on evaluation, future direction, and a final Q\&A session.

\subsection{Agenda}
A tentative schedule of the tutorial is as follows.
\begin{enumerate}
    \item Introduction \textbf{(10 min)}
    \begin{enumerate}
        \item Conversational Systems 
        \item Problem of Data Scarcity
        \item Data Augmentation
    \end{enumerate}
    \item Conversation Augmentation \textbf{(30 min)}
    \begin{enumerate}
        \item Generic Token-level \& Sentence-level Augmentation
        \item Dialogue Data Augmentation    
    \end{enumerate}
    \item Conversation Generation: Open Domain \textbf{(80 min)}
    \begin{enumerate}
        \item Single-turn QA Pair Generation    
        \item Multi-turn Dialogue Generation
        \item Topic-aware Dialogue Agent
    \end{enumerate}
    \item Conversation Generation: Task-oriented \textbf{(40 min)}
    \begin{enumerate}
        \item Schema-guided Generation
        \item Simulator-Agent Interaction
        \item E2E Dataset Creation
    \end{enumerate}
    \item Evaluation \textbf{(10 min)}
    \item Conclusion and Future Direction \textbf{(10 min)}
\end{enumerate}

\subsection{Content}

\paragraph{\textbf{Introduction}}
We start by introducing the audience to the basics of conversational systems, including TOD and ODD systems. We provide an overview of the components and concepts associated with TOD and ODD systems, ensuring that participants can grasp the necessary knowledge to follow the tutorial independently.
We further discuss the data scarcity problem in creating dialogue systems, particularly in low-resource domains and languages and give an introduction to the proposed techniques to tackle this issue. 
Given that dialogue datasets may not be available for all languages and domains, we discuss dialogue generation methods that leverage external resources  such as unstructured text files, knowledge graphs, and Large Language Models (LLMs). 

\paragraph{\textbf{Conversation Augmentation}}
We proceed by providing an overview of existing works in conversation augmentation.
Augmentation techniques have demonstrated effectiveness in various NLP tasks, involving the creation of new samples through modifications of existing ones.
However, augmenting dialogue data requires precision due to the interconnected nature of multi-turn user-bot utterances, presenting additional challenges.
Within augmentation methods, there are two broad categories applicable to text-based tasks: token-based~\cite{Jungiewicz2019MaxLoss, wei2019eda, Miao20Snippext, kobayashi2018CAPR, gao2019SCDA, kumar2020data, jia2016data, andreas2020GECA, yoon2021ssmix, zhang2022treemix} and sentence-based~\cite{Yu18QANet, Bornea21MTLQA, kumar2019DiPS} approaches. These categories involve the replacement of original tokens or sentences with relevant alternatives.
We discuss specific techniques have been proposed to generate new dialogue samples for both TOD~\cite{Yin2020RDA, gao2020PARG, Li2021coco, lai2022CUDADST} and ODD~\cite{Li2019CVAEGAN, zhang2020DlDistill, Ou22CAPT, poddar2022dialaug} systems. These models employ generative models, RL-based models, counterfactual learning, or user dialogue act augmentation and offer new avenues to generate dialogue samples, further enriching the available training data for dialogue systems.

\paragraph{\textbf{Conversation Generation: Open Domain}}
In this part of our tutorial, we focus on the methods available for generating dialogue samples for an ODD system.
The pipeline approach, initially introduced for synthetic QA pair generation~\cite{alberti2019synthetic}, is one way to generate ODD samples. This method consists of four sequential stages: passage selection, answer extraction, question generation, and a subsequent filtering process to maintain quality of generated QA pairs.
Building upon the successful generation of QA pairs~\cite{puri2020training, lewis2021paq, Yang2020DAug, maharana2022grada}, researchers have extended the pipeline approach to generate complete conversation samples, addressing challenges such as sub-passage selection, flow consistency, coreference alignment, and handling different question types~\cite{gao2019CFnet, gu2021chaincqg, nakanishi2019AUCQG, do2022cohsCQG, wu2022dg2, hwang2022multiCQG, kim2022simseek}. However, a major limitation of the pipeline approach is that the conversation's flow is primarily determined by the passage's flow. This means that a passage is initially divided into multiple chunks, and each turn of the conversation is generated based on its corresponding chunk. To achieve more control over the conversation flow, one potential solution involves generating a multi-turn conversation along a path of entities or keywords extracted from a knowledge graph (KG). Based on this idea, various tasks have been defined to connect the initial entity or sentence to the target entity. One such task is the one-turn topic transition, which generates a "bridging" utterance to connect the newly introduced topic to the previous turn's topic~\cite{sevegnani2021otters, gupta2022CRG, kishinami2022TGCP}. Additionally, we introduce target-oriented dialogue systems, where models actively guide conversations towards predefined target topics, ensuring smooth transitions and progress towards the desired targets~\cite{tang2019target, Qin2020DKRN, Zhong2021CKC, yang2022topkg}.
Furthermore, we explore goal-directed dialogue planning strategies that empower the dialogue system to embrace a discourse-level perspective, taking into account the overarching objective of the conversation, with the aim of generating a response that aligns with it~\cite{Xu20KnowHRL, Xu20EGRL, ni2022hitkg, Wang2023COLOR}.

\paragraph{\textbf{Conversation Generation: Task-oriented}}
We next discuss the conversation generation methods for TOD systems. 
In such systems, the primary objective of the dialogue system is to understand the user's intent throughout a multi-turn conversation and subsequently provide relevant suggestions to assist the user in achieving their goal. However, accurately capturing the essential information from the user's utterances to ensure successful task completion requires meticulous attention and domain expertise~\cite{Jinjie2023Recent}.

We begin by introducing schema-guided generation methods~\cite{Shah2018M2M, Rastogi2020SGD}, which leverages self-play models to generate dialogues. The generated dialogues are then annotated and filtered using crowdsourcing techniques.
Another approach focuses on enhancing user simulator models, which simulate user behavior and engage in conversations with dialogue systems to generate dialogues for training and evaluation~\cite{zhao2019LaRL, lin2021TUS, tseng2021DSUSRL, mohapatra2021simulatedchats, wan2022unified, Terragni2023ICLUS}. Improvements in these simulators contribute to overall enhancements in dialogue system performance and their ability to handle diverse user inputs and scenarios.
Lastly, we explore end-to-end approaches that aim to directly generate dialogue without explicitly defining intermediate steps or modules~\cite{yoo2020VHDA, kim2021neuralwoz, mehri2022lad, li2022DIALOGIC, chen2023places}. End-to-end models offer the advantage of encapsulating the entire dialogue generation process within a single model, simplifying both training and inference procedures.

\paragraph{\textbf{Evaluation}}
After discussing dialogue data creation methods, we turn into evaluating the quality of these data.
The evaluation process encompasses two levels: turn-level evaluation and global-level evaluation~\cite{hwang2022CQAGAR, hwang2022multiCQG, Zhong2021CKC, yang2022topkg}. At the turn-level, the system's response is compared to the ground-truth response, and this evaluation primarily relies on automatic metrics. Moving to the global-level evaluation, the aim is to assess the overall conversation quality by considering characteristics such as naturalness, coherence, answerability, and success rate in achieving targets. This evaluation level involves generating conversation samples through interactions between the dialogue system and a user simulator or a human, followed by scoring the entire conversation sample. Alternatively, the generated data can be used to train downstream tasks~\cite{kim2021neuralwoz, mohapatra2021simulatedchats, wan2022unified}, and the resulting improvements in performance can be measured. We thoroughly discuss the advantages and disadvantages of these evaluation methods, considering their suitability for different scenarios.


\paragraph{\textbf{Conclusion and Future Direction}}
We conclude the tutorial with an exploration of open research problems and future directions in the field.

%% file: sections/presenters.tex
\section{Presenter Biography}

\noindent
\textbf{Heydar Soudani} 
is a first-year Ph.D. student at Radboud University's Institute of Computing and Information Sciences (iCIS), where he is being supervised jointly by Faegheh Hasibi and Evangelos Kanoulas.
He holds a Bachelor's degree from Polytechnic of Tehran and a Master's degree from Sharif University of Technology.
His research primarily focuses on conversational systems in low-resource domains and languages. Specifically, he is dedicated to the development of knowledge-grounded models that generate synthetic multi-turn conversation data.

\vspace{0.5em}
\noindent
\textbf{Faegheh Hasibi} is an assistant professor of information retrieval at the Institute of Computing and Information Sciences (iCIS) at Radboud University. Her research interests are at the intersection of Information Retrieval and Natural Language Processing, with a particular emphasis on conversational AI and semantic search systems. She explores various aspects, including knowledge-grounded conversational search, entity linking and retrieval, and the utilization of knowledge graphs for semantic search tasks. Her contributions to the field are published in renowned international conferences such as SIGIR, CIKM, COLING, and ICTIR and have been recognized by awards at the SIGIR and ICTIR conferences.
She has given mutilple invited talks and has extensive experience as a lecturer. 

\vspace{0.5em}
\noindent
\textbf{Evangelos Kanoulas} is a full professor of computer science at the University of Amsterdam, leading the Information Retrieval Lab at the Informatics Institute. His research lies in developing evaluation methods and algorithms for search, and recommendation, with a focus on learning robust models of language that can be used to understand noisy human language, retrieve textual data from large corpora, generate faithful and factual text, and converse with the user. Prior to joining the University of Amsterdam, he was a research scientist at Google and a Marie Curie fellow at the University of Sheffield. His research has been published at SIGIR, CIKM, WWW, WSDM, EMNLP, ACL, and other venues in the fields of IR and NLP. He has proposed and organized numerous search benchmarking competitions as part of the Text Retrieval Conference (TREC) and the Conference and Labs of the Evaluation Forum (CLEF). Furthermore, he is a member of the Ellis society (https://ellis.eu/).